%% file: main.tex
\documentclass[letterpaper, 10 pt, conference]{ieeeconf} 
\usepackage{graphicx} 
\usepackage[table]{xcolor}% http://ctan.org/pkg/xcolor
\usepackage{multirow}
\usepackage{subfigure} 
\usepackage{hyperref}                                   %hyperlink
\hypersetup{
    colorlinks=true,
    linkcolor=blue,
    filecolor=magenta,      
    urlcolor=black,
}
\def\BibTeX{{\rm B\kern-.05em{\sc i\kern-.025em b}\kern-.08em
    T\kern-.1667em\lower.7ex\hbox{E}\kern-.125emX}}

\usepackage[normalem]{ulem}
\useunder{\uline}{\ul}{}
\IEEEoverridecommandlockouts                              % This command is only needed if 
\title{\LARGE \bf
Delivering Cognitive Behavioral Therapy Using A Conversational Social Robot}

\author{Francesca Dino$^{1}$, Rohola Zandie$^{2}$, Hojjat Abdollahi$^{2}$, Sarah Schoeder$^{3}$ and Mohammad H. Mahoor$^{2}$% <-this % stops a space
\thanks{$^{1}$Department of Psychology, University of Denver, 
        {\tt\small francesca.dino@du.edu}}%
\thanks{$^{2}$Department of Electrical and Computer Engineering, University of Denver, Colorado, USA.
        {\tt\small (roholazandie,habdolla,mmahoor)@du.edu}}%
\thanks{$^{2}$Eaton Senior Community, Lakewood, CO. 
        {\tt\small sschoeder@eatonsenior.org}}%\textbf{}
}

\begin{document}

\maketitle
\thispagestyle{empty}
\pagestyle{empty}

%%%%%%%%%%%%%%%%%%%%%%%%%%%%%%%%%%%%%%%%%%%%%%%%%%%%%%%%%%%%%%%%%%%%%%%%%%%%%%%%
\begin{abstract}
Social robots are becoming an integrated part of our daily life due to their ability to provide companionship and entertainment. A subfield of robotics, Socially Assistive Robotics (SAR), is particularly suitable for expanding these benefits into the healthcare setting because of its unique ability to provide cognitive, social, and emotional support. This paper presents our recent research on developing SAR by evaluating the ability of a life-like conversational social robot, called Ryan, to administer internet-delivered cognitive behavioral therapy (iCBT) to older adults with depression. For Ryan to administer the therapy, we developed a dialogue-management system, called Program-R. Using an accredited CBT manual for the treatment of depression, we created seven hour-long iCBT dialogues and integrated them into Program-R using Artificial Intelligence Markup Language (AIML). To assess the effectiveness of Robot-based iCBT and users' likability of our approach, we conducted an HRI study with a cohort of elderly people with mild-to-moderate depression over a period of four weeks. Quantitative analyses of participant's spoken responses (e.g. word count and sentiment analysis), face-scale mood scores, and exit surveys, strongly support the notion robot-based iCBT is a viable alternative to traditional human-delivered therapy.
\end{abstract}

%%%%%%%%%%%%%%%%%%%%%%%%%%%%%%%%%%%%%%%%%%%%%%%%%%%%%%%%%%%%%%%%%%%%%%%%%%%%%%%%
\section{INTRODUCTION}
\begin{figure}[]
	\centering
	\includegraphics[width=8cm, trim={0cm 5.2cm 0cm 3.1cm},clip]{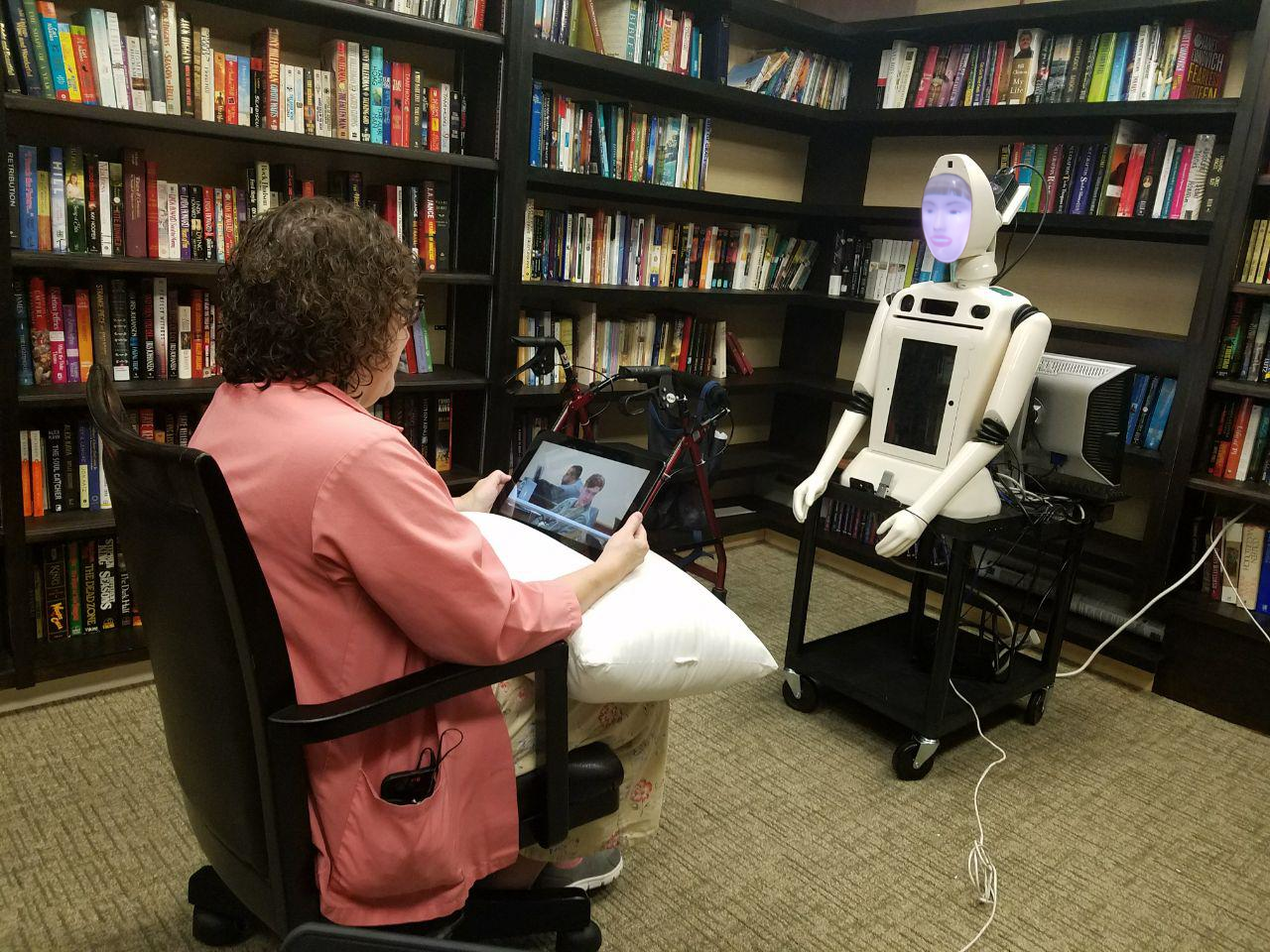}
	\caption{An example of Ryan the CompanionBot interacting with a user in the experimental setup.}
	\label{fig:experiment}
	\vspace{-5mm}
\end{figure}

Recent exciting technological advances and research in the field of robotics has led to commercially available personal robots ranging from industrial machines to human-like androids.  These robots, such as Ryan (see Fig. \ref{fig:experiment}), a social robot capable of complex human interaction \cite{dreamfacetech}, possess a wide range of functionalities that allow them to be useful in society as sources of entertainment, platforms for research, and tools for medical professionals. Robots are even beginning to find their own place in our homes as the continual improvement of their verbal and nonverbal socioemotional capabilities enriches human-robot interaction (HRI).

A sub-field in robotics, Socially Assistive Robotics (SAR), especially strengthens HRI due to its unique ability to provide emotional, cognitive, and social support \cite{feil2005defining}. Effective human-robot communication, made possible by advances in natural language processing and machine comprehension, allows intelligent conversational systems to provide commercial and customer service in healthcare settings. Therapy applications are of particular interest as natural language generation in modern dialog systems creates conversation that appears fluid and natural to the users. This flexible conversation necessary for therapy can be achieved by using natural language understanding to extract each user's different expressions and matching their linguistic level. Particularly, continual recording and extraction of patient mood, sentiment, sentence length, etc. using an information extraction algorithm makes it possible for SAR to help develop more comprehensive treatments. 

One such practical implication for robots and advanced dialog systems is the treatment of depression. Late-life depression (LLD), a type of depression occuring in individuals aged 60 and older, is a serious medical illness impacting 8 to 16\% of older adults \cite{doi:10.1002/gps.4325}. This is especially concerning because depression in older adults is associated with worsening of existing physical illness, cognitive impairment, and an increased risk of suicide \cite{2011-09149-00820110601}. Effective treatment of LLD has been achieved with a form of psychotherapy called cognitive behavioral therapy (CBT) \cite{beck2011cognitive}. When CBT is delivered through technology-based platforms, it is called internet-delivered cognitive behavioral therapy (iCBT) \cite{ 2014-53983-00120150301}. Past research has suggested iCBT is beneficial and can be just as effective as traditional CBT in providing therapy to older adults \cite{2013-12181-01220130201, 2013-32985-00120131201}.  

Due to financial burden, mobility limitations, and the stigma associated with mental illness, older adults suffering from depression often do not seek out and receive the help they so desperately need \cite{2013-12181-01220130201}. To address some of these barriers and develop novel approaches for the treatment of mental illness, special consideration has been put into developing companion and service robots with abilities to assist humans at a socio-emotional level. SAR is especially appealing for older adults as it can be used in a way that allows for patients to remain in their preferred environment and access effective, affordable treatment \cite{RABBITT201535}. This research combines work on iCBT, advanced dialog systems, and SAR to further develop robotics technology to administer CBT to older adults suffering from depression. 

The remaining sections of this paper are as follows. Section \ref{relatedwork} reviews current literature describing advancements in dialogue systems and social robotics. Section \ref{robotbasediCBT} presents the robotic platform, dialogue manager system, and session dialogues developed in this research. An evaluation of this technology with human subjects as well as the research methodology can be seen in Section \ref{humansubjectevaluation}. Results and analysis of the data is explained in Section \ref{results}. Section \ref{discussion} discusses the importance of the results, and finally Section \ref{conclusion} concludes the paper and offers suggestions for future research.

\section{RELATED WORK}
\label{relatedwork}

Continual refinement of emotion recognition and natural language processing techniques has allowed for chatbots and dialogue systems to be successfully used in therapy and counseling settings. One study attempted to redefine emotion recognition by creating an unobtrusive system to measure emotions with the use of smartphones \cite{lee2012towards}. This system eliminated the use for expensive and clunky sensors, as well as demonstrated high accuracy in classifying user emotions into categories. In \cite{oh2017chatbot}, a chatbot was developed that used natural language processing techniques to recognize emotion and respond accordingly, but unlike the current study, the chatbot focused on everyday conversation and had no predefined counseling schema. Another experiment created a chatbot with emotional capabilities, however for sentence generation, they used general knowledge bases without the specific vernacular needed for counseling \cite{lee2017chatbot}. The closest research to the work described in this paper is \cite{bickmore2011reusable}, which is implemented using an OWL ontology of health behavior concepts. Even though this approach is extensive and increases complexity, it is not flexible to different user inputs and is very hard to extend. 

The field of SAR has taken these dialogue systems one step further by exploring the use of a physical robot to deliver conversation to the user.  Moro \textit{et al.} explored human-like robots as companions to seniors and found the more human in appearance the robot was, the more engagement and positive effect it had \cite{2018-40296-00120180810}. The ability to communicate with humans in a socially appropriate manner enables these robots to be sources of medical treatment in addition to just social companions. Animal robotic companions like PARO, an advanced interactive robot designed to look like a seal, are popular in this line of work \cite{paro}. PARO has been found to produce benefits similar to that of animal therapy such as reducing patient stress, improving motivation, and increasing patient socialization \cite{doi:10.1111/jnu.12423}. Wada \textit{et al.} used PARO to study the effect of social robots on residents of a senior care center and found that the residents maintained a lower stress level and established rapport with the robot~\cite{wada2003effects}. Kargar and Mahoor used an animatronic, artificially intelligent social robotic bear named eBear with older adults with moderate depression and concluded that it helped improve their mood~\cite{kargar2017pilot}. The humanoid robot NAO \cite{nao} has been used to combat cognitive decline and depression through various methods such as making jokes, dancing, playing music, and most importantly, conversing with the patients \cite{2018-05757-00120180207}.

\section{Robot-Based iCBT} 
\label{robotbasediCBT}

\subsection{Ryan}
The robotic platform used in this research is Ryan Companionbot~\cite{dreamfacetech} shown in Fig \ref{fig:experiment}. Ryan is a social robot created by DreamFace Technologies, LLC. for face-to-face communication with individuals in different social, educational, and therapeutic contexts. This robot has been used in studies with older adults~\cite{abdollahi2017pilot}, children with autism~\cite{askari2018}, and other social robotics research~\cite{Mollahosseini2018, mollahosseini2018role}. Ryan has an emotive and expressive animation-based face with accurate visual speech and can communicate through spoken dialogue. The robot's face uses a rear projection method due to the difficulty and expense of using actuators to build a natural face capable of showing visual speech and emotions. Having the capability of showing subtle communication features makes Ryan appealing to the elderly and a suitable platform for this HRI study. 
 
There is a touch-screen tablet on Ryan's chest that can show videos or images to the user and receive the user's input. An external (detached) tablet was used in this research to allow the subjects to interact with Ryan while sitting a comfortable distance from the robot. A remote controller software, nicknamed Wizard of Oz (WOZ), was developed for Ryan so that in the case an intervention was necessary, the researchers were able to control the flow of the experiment by talking through the robot to the users without breaking the perception of autonomy of the robot.

The robot is equipped with a dialog manager called ProgramR (explained in next section). The user input is sent to the dialog manager as a string of text and the proper response is then sent back to the robot software. Ryan uses a Text-To-Speech engine to convert the response into audio and phonetic timing for the animation. Ryan then uses this information to talk to the subject. If the response from the dialog manager refers to an image or a video clip, the proper media file will be displayed on the tablet screen. The software structure of the robot is illustrated in Fig. \ref{fig:programr}. The work by Abdollahi \textit{et al.}~\cite{abdollahi2017pilot} contains more details on the hardware and software of the robot.

\subsection{Program-R}

To deliver the iCBT for this study, we developed an AIML-based dialogue manager system called Program-R. Program-R is a forked project from Program-Y \cite{program-y} with several modifications to fit the purpose of this study and create a seamless interaction between Ryan and the subjects. 

AIML is an XML-based language for writing conversations. Among AIML objects, the following tags are worth citing: ``category," ``pattern," ``template," and ``that". The ``category" tag is the basic unit of dialogue that includes other predefined tags. The ``pattern" tag defines a possible user input and the ``template" tag is a certain response from the dialogue manager. To ensure continuity between the dialogues, the ``that" tag is used to connect different dialogues with the last question asked by the dialogue system. 
% Figure \ref{fig:programr} illustrates a unit of dialogue in our repository.

In order to deliver a more interactive user experience, we introduced the ``robot" tag. The ``robot" tag is the information that is used by Ryan to enhance HRI with the use of multimedia. Inside the ``robot" tag there is an ``options" tag that lists the possible answers to the questions at the end of the current dialogue response. Moreover, the ``image" and ``video" tags provide Ryan with specific multimedia information to be presented to the user by Ryan.

When all the dialogues are connected to each other with the ``that" tag, a graph structure is created, rather than a tree structure, because the branches of the flow of conversation can merge back again. The control structure of conversations is defined as session frames in a finite-state automaton. The conversation involves a few slot fillings, such as, the name of the user, place of birth, and answers to questions that shape the conversation pattern. All of this user information is saved in a database for use in future sessions.
% \begin{figure} 
%   \includegraphics[width=\textwidth]{program-r.png}
% \end{figure}

\begin{figure*}[]
                \centering
                \includegraphics[width=\textwidth]{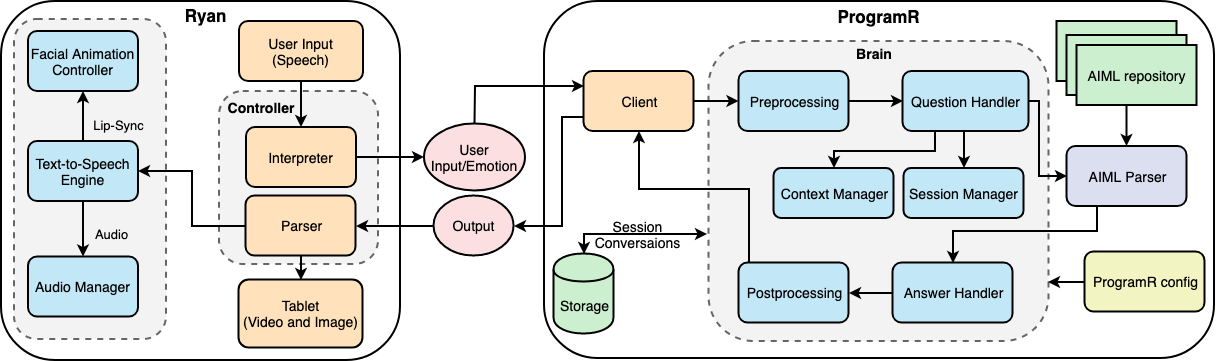}
                \caption{The proposed system diagram.}
                \label{fig:programr}
                \vspace{-1em}
\end{figure*}

Unlike most dialogue managers, Program-R is an active system, starting the conversation and asking questions of the user, rather than the user initiating conversation. Fig \ref{fig:programr} demonstrates the architecture of our dialogue system.  Program-R communicates with Ryan through a Representational State Transfer RESTful API \cite{restfulapi}. The user input from the speech to text component is received by the proper Client and then sent to Brain. Brain is the core module which handles several key tasks. It takes care of communicating with storage for cases of resuming the interrupted conversation and connects with the AIML parser to resolve the answers with the consultation of the AIML repository. In Brain, the input will be preprocessed to remove punctuation, normalize the text, and segment the sentences. In the next step, Question Handler adds the context and session data to the input. Context Manager handles the context in which the conversation is happening. For example, two different questions can have yes/no answers, but without knowing the context in which the conversation is happening, responding to them is impossible. In these cases, Context Manager helps to respond properly. In addition, the Context Manager provides custom responses to different users based on the information that was recorded previously for that specific person. Meanwhile, Brain saves all the conversations and session data in the form of explicit (i.e. name or place of birth) and implicit (i.e. mood) user information in the database. Finally, the postprocessing (i.e. formatting numbers, remove redundant spaces and html tags, etc.) will be performed on the answer and the result will be sent to Ryan. 

\begin{figure}[tb]
    %\centering
    \includegraphics[width=8.8cm, clip]{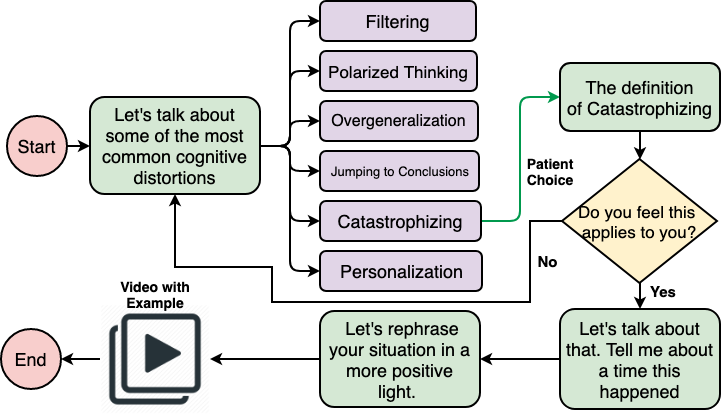}
    \caption{A sample dialogue of a conversation about cognitive distortions between Ryan and the user. Users have the chance to choose a distortion that relates to them, discuss it, and learn how to cognitively reappraise the situation with the aid of a visual example.}
    \label{fig:flowchart}
    \vspace{-1em}
\end{figure}

Due to the fact the dialogue manager works based on an automaton, there is a risk that the user will drift away from the conversation flow by answers that are irrelevant to the current question. In these situations, the dialogue manager can ask the question again with a different format or give control to the WOZ to handle the situation.

Unlike other dialog systems used for counseling and mental healthcare, the proposed approach can interact with patients in more diverse ways with the help of images, music, videos, and the presence of a robot.

\subsection{Session Dialogues} 
Therapy sessions were formatted using Artificial Intelligence Markup Language (AIML). The AIML repository contained 7 AIML files that were organized in a session-based manner to follow the structure of iCBT as described in an individual CBT therapy plan for depression \cite{munoz2000individual}. The seven treatment sessions were spread out over the span of four weeks and were broken down into three key points: how thoughts affect mood (sessions 2 and 3), how activities affect mood (sessions 4 and 5), and how people affect mood (sessions 6 and 7). Session 1 consisted of a general introduction and allowed for participants to familiarize themselves with Ryan. An example of a dialogue between Ryan and a user can be seen in Fig \ref{fig:flowchart}. In total, there were 165 categories, 23 robot tags, and 27 additional media (10 pictures, 13 videos, and 4 music files). All pictures, videos, and music were educational or therapy-driven in purpose. Therapy sessions were executed by Program-R, the dialogue manager developed in this research.

\section{Human Subject Evaluation}
\label{humansubjectevaluation}

\hbox{}\subsection{HRI Study Design}
To evaluate the feasibility of using a conversational social robot to deliver iCBT, participants were recruited to undergo the therapy described earlier in this paper. Several mental health evaluation tests were conducted prior to the therapy sessions and after the conclusion of the treatment period for comparison. Mental health examinations included the Saint Louis University Mental Status Examination (SLUMS) \cite{FELICIANO2013623} to assess cognitive deficits, the Patient Health Questionnaire-9 Item (PHQ-9) \cite{doi:10.1046/j.1525-1497.2001.016009606.x} and Geriatric Depression Scale (GDS) \cite{doi.org/10.1300/J018v05n01_09} to observe depression symptoms, as well as a Face Scale Mood Evaluation \cite{doi:10.1002/art.1780290714} to gain a day-to-day sense of participant mood.

Administration of the treatment took place at the library within the senior living facility due to its privacy and shelter from outside noise. Participants were seated one-on-one with Ryan and visual and audio of each session was recorded for transcription and data analysis purposes. Twice a week for about an hour, participants met with Ryan at their scheduled time to go through the therapy dialogues. Face scale scores were gathered at the start and end of each session for each participant. Following the conclusion of the last therapy session, an exit interview was conducted to gather subject feedback and evaluate bot functionality.
\input{demographics.tex}

\subsection{Participants}
The four participants for this study were chosen from Eaton Senior Communities, an independent living facility located in Lakewood, CO. Each participant selected was over the age of 60, showed at worst mild cognitive impairment or no impairment at all, and scored within the range of mild to severe depression on the assessment tools administered (shown in Table \ref{tab:demographics}).  Additionally, each subject was selected to have availability for two, one-hour sessions twice a week. Prior to participating, subjects were briefed fully on the study design and consented to their involvement, with the proper Institutional Review Board (IRB) approvals for human-subjects in place.

\input{exitsurvey.tex}
\section{RESULTS}

\subsection{Natural Language Analysis}

\begin{figure}[b]
    %\centering
    \includegraphics[width=8.8cm, height=3cm, trim={0 0.9cm .0cm 0.0cm},clip]{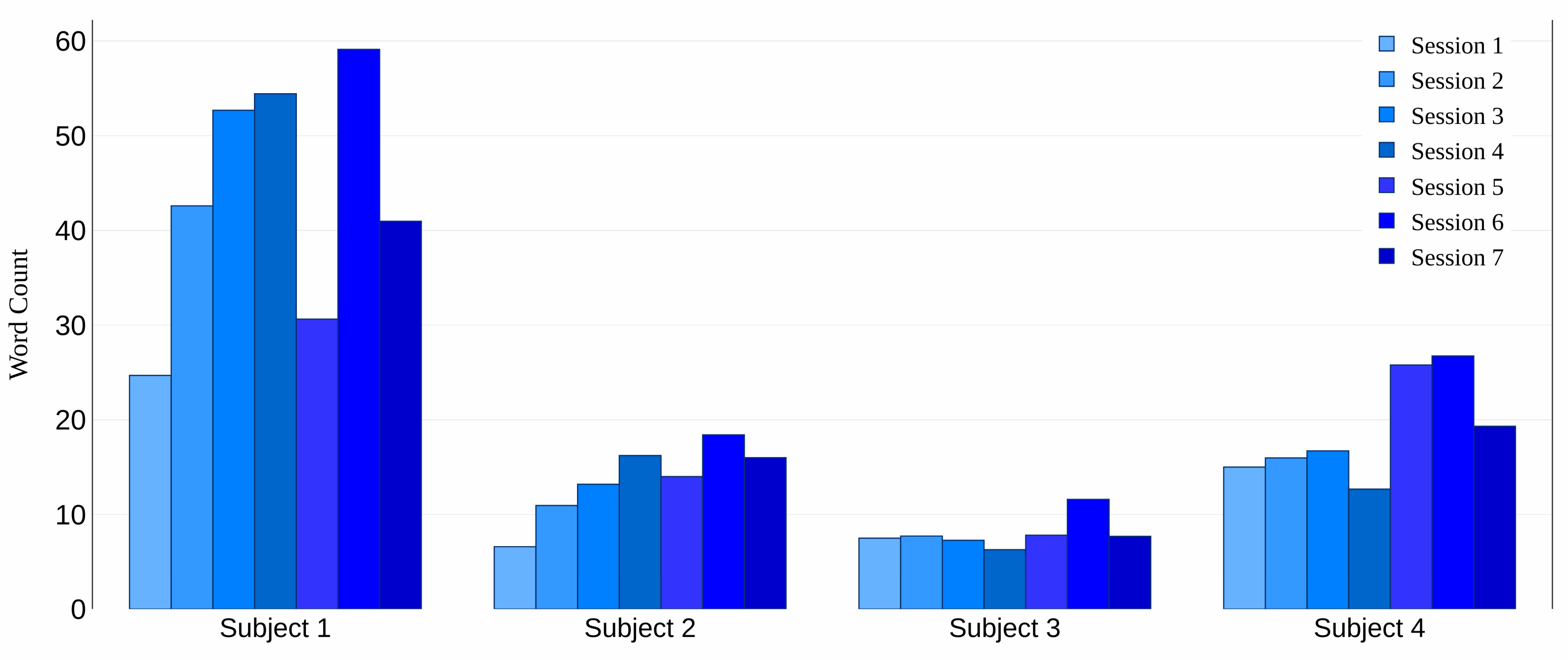}
    \caption{Word counts for each subject over the seven sessions.}
    \label{fig:wordcount}
    \vspace{-1em}
\end{figure}

Natural language analysis was used to evaluate the degree of subject involvement throughout the sessions. One measure of involvement is the average response length by the user to questions asked by the robot. To calculate the average response length, the user responses were tokenized and the average length of tokens per session was calculated. Fig \ref{fig:wordcount} demonstrates the rough increase of word count of each subject as the sessions progressed. Excluding the last session-a closing session involving a wrap up of the whole study-an increase in average sentence length can be seen in almost all of the participants. In order to decrease the bias on this evaluation, the sessions were designed to have almost the same number of questions. For example, in session 6, which shows the most involvement in all the participants, 18 categories were used whereas the first session has 23 categories. There is also another phenomenon that can be seen in this plot: some participants tended to give longer responses to the questions than others. Despite this, the fact that individual participants talked longer as the sessions progressed still holds.

\begin{figure*}[t!]%
\small
\centering

\includegraphics[height=3in]{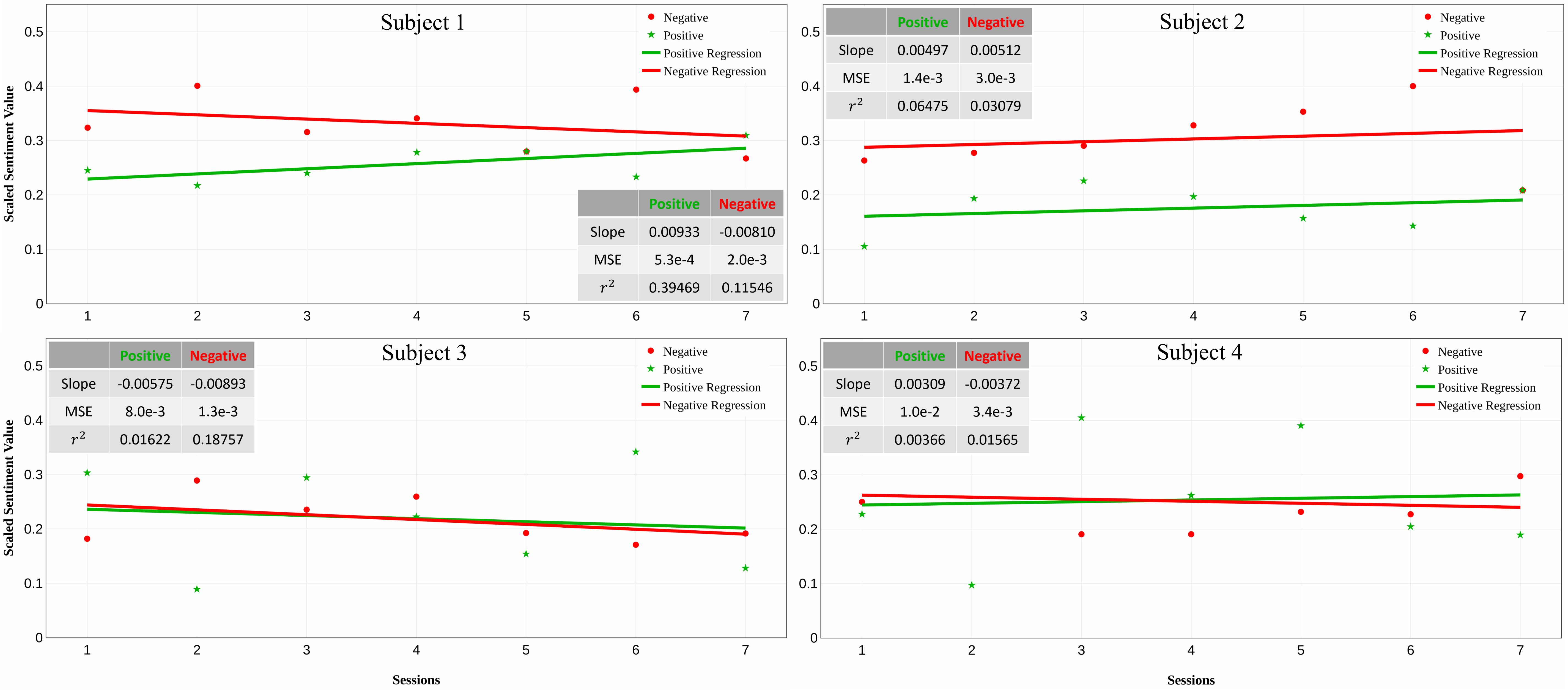}

\caption[]{Scaled sentiment values (positive and negative) and their linear regression for four subjects and seven sessions. The table in each figure shows the slope, mean squared error and variance score for positive and negative sentiment regressions.
}%
\label{fig:sentiment}%
\end{figure*}

Another measure of subject involvement is sentiments over time. Sentiment analysis is a technique to evaluate the positive, negative, and neutral sentiments at a sentence level. CoreNLP \cite{manning-EtAl:2014:P14-5} was used to measure the sentiments. First, transcriptions of each sentence spoken by the user were segmented and then the sentiments of each sentence were computed. Stanford CoreNLP has two more categories for sentiments: ``verypositive'' and ``verynegative.'' For the purposes of this research, these categories were considered as positive and negative respectively, because they occur very infrequently compared to the other three sentiments. The number of positive, negative, and neutral sentiments are scaled such that their overall summation becomes 1. In Fig \ref{fig:sentiment}, the neutral contribution in each session was excluded due to the fact that they do not show any mood from the user. The results of the sentiment analysis for each subject was unique. Fig \ref{fig:sentiment} shows an increase in positive sentiments and a decrease in negative sentiments for subjects 1 and 4. For subject 2, the positive sentiment increased whereas the negative sentiment increased with the same rate and for subject 3, the negative sentiment decreased faster than the positive sentiment. The fluctuation in sentiment value for the last two subjects was higher. Overall, the sentiment analysis shows improvement for two subjects and more inconsistent results for the other two.

\label{results}

% \begin{figure*}[t!]%
% \small
% \centering
% \subfigure[][]{%
%     \label{fig:ex3-a}%
%     \includegraphics[height=1.85in, width=3.4in]{figures/1111.png}
% }%
% \subfigure[][]{%
%     \label{fig:ex3-b}%
%     \includegraphics[height=1.85in, width=3.4in]{figures/2222.png}
% } 
% \subfigure[][]{%
%     \label{fig:ex3-c}%
%     \includegraphics[height=1.85in, width=3.4in]{figures/3333.png}
% }
% \subfigure[][]{%
%     \label{fig:ex4-d}%
%     \includegraphics[height=1.85in, width=3.4in]{figures/4444.png}
% }

% \caption[]{Scaled sentiment values (positive and negative) and their linear regression for four subjects and seven sessions. The table in each figure shows the slope, mean squared error and variance score for positive and negative sentiment regressions.
% \vspace{-3em}
% \subref{fig:ex3-a} Subject 1;
% \subref{fig:ex3-b} Subject 2;
% \subref{fig:ex3-c} Subject 3; and,
% \subref{fig:ex4-d} Subject 4.}%
% \label{fig:sentiment}%
% \end{figure*}

%feel free to move this wherever looks the best
%face scale results
\begin{figure}[tb]
    \centering
    \includegraphics[width=8cm,  trim={0 0 0 0.0cm},clip]{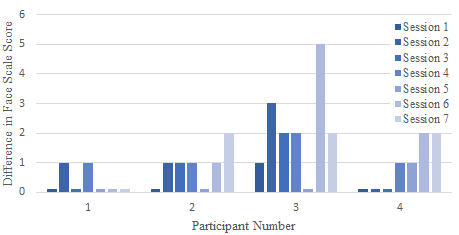}
    \caption{Results of the Face Scale measures indicated by the difference between the before and after scores of each session.}    \label{fig:faceScale}
    \vspace{-2em}%I changed this from -1em to make everything fit
\end{figure}

\subsection{Symptom Outcomes}
As seen in Table \ref{tab:demographics}, pre- and post-therapy scores for the mental health examinations demonstrated improvement in three out of the four subjects for the PHQ-9. GDS scores improved for two out of the four subjects, stayed the same for one out of the four subjects, and worsened in one of the four subjects. When looking at the face scale mood evaluation seen in Fig \ref{fig:faceScale}, it was found that all subjects either stayed the same after treatment or improved. Subject 3 remained more variable throughout the treatment period, with session 6 demonstrating an unusually high improvement in score. Subject 1 was the most consistent in scores, either not improving or only improving by a small amount. Subjects 2 and 4 had more consistent increases in mood throughout the duration of the therapy.

Upon completion of the study, the participants were asked to complete an exit survey. As shown in Table \ref{tab:questionnaire}, participants rated both the interaction with Ryan and the therapy very positively. There was also minimal deviation in scores for each question. Questions three and seven received the lowest average score with a rating of four. The high values for Cronbach's alpha suggest the scale for the exit survey had consistency and reliability.

\section{DISCUSSION}
\label{discussion}
This research demonstrated promising results about the positive impact of using a social robot to administer iCBT to older adults with depression. Natural language analysis demonstrated that as the individual subjects progressed through the sessions, their average sentence length increased. This is likely due to elevated feelings of comfort and enjoyment with Ryan. It could also suggest a decrease in symptom severity as less distress from depression symptoms could allow the participants to engage more. This is further supported by the fact that sessions with less programmed dialogue (i.e. Session 6) had increased sentence length, despite having less material to cover.  
Sentiment analysis also suggested that patient mood improved throughout the course of treatment. For subjects 1 and 4 specifically, an increase in positive sentiments and a decrease in negative sentiments indicates content of subject speech became less pessimistic as the sessions progressed. Subjects 2 and 3 had more fluctuation between sessions, but still show the same trends. This is a likely sign that depression symptoms were diminishing as the subjects increasingly included optimistic language in their speech. 

The day-to-day information gathered by the mood scale provides further evidence of subject improvement. Not only did subjects either stay the same or feel better after interacting with Ryan, the amount of improvement for many of the subjects appeared to increase as the sessions continued. This suggests that the subjects felt better on a daily basis and over the course of the entire therapy duration. These results are similar to conclusions in other research discussed regarding a reduction of depressive symptoms following social robot intervention \cite{doi:10.1111/jnu.12423}.

Mental health evaluations produced inconsistent results. Displayed in Table \ref{tab:demographics}, three out of the four subjects showed improved PHQ-9 scores and two out of the four subjects showed improvement in GDS scores. Specifically, with the GDS results, subject 4 did not show a change in scores and the score for subject 2 increased. The subjects that either did not show improved scores or showed worsened scores were not the same for both tests. Differences between which subject did not improve between the two tests could be explained by individual characteristics and the unique approaches each test takes to measuring symptoms of depression. For example, the PHQ-9 is a longer, more comprehensive test, and the GDS tends to be a more surface level examination, which may reflect a difference in the scores.

The exit survey results demonstrate that overall, patients were satisfied with Ryan and the therapy delivered by her. As seen in Table \ref{tab:questionnaire}, the lowest average score on a question was a four, meaning most questions were rated very positively. The low amount of deviation for each score suggests that the subjects tended to all feel very similarly about what each question was asking. Specifically, the highest rated questions suggested the participants enjoyed interacting with the robot and found the information presented to them in the therapy highly valuable. 

Open ended survey questions allowed the participants to more freely express their opinion. The subjects conveyed that they enjoyed their time with Ryan and were amazed by her functionality. As said by one participant, ``Her [Ryan's] responses at the beginning set the mood. How special to have someone begin my day with a smile and happy voice!" Positive subject feedback in this setting is especially important as previous studies \cite{mollahosseini2018role, Mollahosseini2018} only took place in a laboratory environment. Now it is clear Ryan can also be successful in delivering cognitive-based therapeutic conversations with elderly human subjects outside of the lab. Criticism of the study suggested session length could be extended and that Ryan could improve her explanation on activities. These are important considerations to address going forward.

It is important to note that for several reasons, it cannot be fully proven that iCBT was the only factor in reducing depression scores. The sample size in this experiment was too small to allow for generalization of the results. Additionally, confounding variables, such as the excitement of participating in an experiment, may have had a factor in lifting the participants' moods. Even if the therapy was not the only factor involved, the results of this study are still important in the fact they demonstrate that even if a robot does not yet fully replace a human therapist, the use of SAR in delivering therapy is a viable alternative to assist those suffering from depression.

\section{CONCLUSIONS AND FUTURE WORK}
\label{conclusion}
This research aimed to provide insight into the feasibility and effectiveness of iCBT administered using SAR as a potential treatment for depression in older adults. To test this, we designed and implemented Program-R, a dialog system that can interact with users using natural language processing techniques through the socially assistive robot Ryan. Human subject testing involved Ryan interacting with four subjects to complete seven sessions of therapy. Based on results gathered after the administration of treatment, the use of SAR in delivering therapy may not yet replace a human therapist, but can provide a viable alternative. Future research should be performed with a larger sample size and over a longer length of time to gather more conclusive results. Hopefully this work will expand upon the use of SAR in the healthcare system and pave the way for more efficient and accessible treatment for those suffering from depression.



\section{ACKNOWLEDGMENT}
This work was partially supported by grant CNS-1427872 from the National Science Foundation. The researchers would like to thank Victoria Miskolci for her help with audio transcriptions.

% \begin{thebibliography}{99}

% \end{thebibliography}

\bibliographystyle{IEEEtran}
\bibliography{IEEEabrv}

\end{document}

%% file: demographics.tex
\begin{table}[b]
\caption{Age, gender, SLUMS score, PHQ-9 score (pre and post study), and GDS score (pre and post study) for each of the subjects.}
\label{tab:demographics}
\begin{center}
\begin{tabular}{ |c|c|c|c|c|c|c| }
\hline
 Sbj & Age/Gender & SLUMS & \multicolumn{2}{c|}{PHQ-9 score}& \multicolumn{2}{c|}{GDS score} \\\cline{4-5} \cline{6-7}
         &        & score & pre & post & pre & post \\
\hline
 1 & 80/F & 23 & 14 & 9 & 13 & 9 \\  
\hline
 2 & 93/M & 20 & 16 & 11 & 15 & 17 \\  
\hline
 3 & 62/F & 27 & 7 & 11 & 13 & 11 \\
\hline
 4 & 69/F & 24 & 13 & 7 & 9 & 9 \\    
\hline
\end{tabular}
\end{center}
\end{table}

%% file: exitsurvey.tex
% \usepackage{multirow}
\begin{table*}[!thb]
\centering
\caption{The questions and mean rank of the exit survey evaluating users' likability and acceptance of interacting with Ryan and the \MakeLowercase{i}CBT module (1-strongly disagree, 5-strongly agree)}
\label{tab:questionnaire}
\begin{tabular}{l|l|c|c|}
\cline{2-4}
                                                                                                                                                 & Question                                                                                                                                  & \begin{tabular}[c]{@{}c@{}} Avg. Score $\pm$ (STD)\end{tabular}   & \begin{tabular}[c]{@{}c@{}} Cronbach's  alpha\end{tabular}        \\ \hline
\multicolumn{1}{|l|}{\multirow{15}{*}{\begin{tabular}[c]{@{}l@{}}Evaluation of \\ Robot Interaction\end{tabular}}} & Q1. I enjoyed interacting with the robot.                                                                                                 & 5.00 $\pm$ 0.00
                      & \multirow{15}{*}{0.88} \\ \cline{2-3}
\multicolumn{1}{|l|}{}                                                                                                                           & Q2. The conversation with the robot was interesting.                                                                                      & 4.75 $\pm$ 0.50
                      &                       \\ \cline{2-3}
\multicolumn{1}{|l|}{}                                                                                                                           & Q3. Learning to interact with the robot was easy.                                                                                         & 4.00 $\pm$ 0.82
                      &                       \\ \cline{2-3}
\multicolumn{1}{|l|}{}                                                                                                                           & Q4. Talking with the robot was like talking to a person.                                                                                  & 4.50 $\pm$ 1.00
                      &                       \\ \cline{2-3}
\multicolumn{1}{|l|}{}                                                                                                                           & Q5. The robot was intelligent.                                                                                                            & 4.75 $\pm$ 0.50
                      &                       \\ \cline{2-3}
\multicolumn{1}{|l|}{}                                                                                                                           & Q6. I feel happier when I was in the company of the robot.                                                                                & 4.50 $\pm$ 0.58
                      &                       \\ \cline{2-3}
\multicolumn{1}{|l|}{}                                                                                                                           & Q7. The robot was acting natural.                                                                                                         & 4.00 $\pm$ 1.15
                      &                       \\ \cline{2-3}
\multicolumn{1}{|l|}{}                                                                                                                           & Q8. The robot encouraged me to talk more.                                                                                                 & 4.75 $\pm$ 0.50
                      &                       \\ \cline{2-3}
\multicolumn{1}{|l|}{}                                                                                                                           & Q9. I feel less depressed after talking to the robot.                                                                                     & 4.50 $\pm$ 1.00
                      &                       \\ \cline{2-3}
\multicolumn{1}{|l|}{}                                                                                                                           & Q10. The robot encouraged me to be more active.                                                                                           & 4.75 $\pm$ 0.50
                      &                       \\ \cline{2-3}
\multicolumn{1}{|l|}{}                                                                                                                           & Q11. I would like to interact with this robot again.                                                                                      & 4.25 $\pm$ 0.96
                      &                       \\ \cline{2-3}
\multicolumn{1}{|l|}{}                                                                                                                           & \begin{tabular}[c]{@{}l@{}} Q12.  I enjoyed using the robot at the end of \\ the month as much as I enjoyed it in the beginning of the study.\end{tabular}                             & 5.00 $\pm$ 0.00
                      &                       \\ \cline{2-3}
\multicolumn{1}{|l|}{}                                                                                                                           & Q13.  I enjoyed the robot playing music for me.                                                                                           & 4.75 $\pm$ 0.50
                      &                       \\ \cline{2-3}
\multicolumn{1}{|l|}{}                                                                                                                           & Q14. I enjoyed the robot playing videos for me.                                                                                           & 4.75 $\pm$ 0.50
                      &                       \\ \cline{2-3}
\multicolumn{1}{|l|}{}                                                                                                                           & \begin{tabular}[c]{@{}l@{}} Q15. The videos played by the robot were effective and helpful \\me either learn something new or affected my life style in a positive way.\end{tabular} & 4.75 $\pm$ 0.50
                      &                       \\ \hline
\hline

\multicolumn{1}{|l|}{\multirow{8}{*}{\begin{tabular}[c]{@{}l@{}}Evaluation of \\ iCBT Module\end{tabular}}} & Q16. I enjoyed the structure of the therapy.                                                                                                & 5.00 $\pm$ 0.00
                      & \multirow{8}{*}{0.76} \\ \cline{2-3}
\multicolumn{1}{|l|}{}                                                                                                                           & Q17. The therapy was organized and made sense.                                                                                      & 5.00 $\pm$ 0.00
                      &                       \\ \cline{2-3}
\multicolumn{1}{|l|}{}                                                                                                                           & Q18. The therapy sessions improved my mood and made me feel happier.                                                                                         & 4.50 $\pm$ 0.58
                      &                       \\ \cline{2-3}
\multicolumn{1}{|l|}{}                                                                                                                           & Q19. I feel like I learned a lot from sessions with the robot.                                                                                  & 5.00 $\pm$ 0.00
                      &                       \\ \cline{2-3}
\multicolumn{1}{|l|}{}                                                                                                                           & Q20. The information presented to me was valuable for my everyday life.                                                                                                          & 5.00 $\pm$ 0.00
                      &                       \\ \cline{2-3}
\multicolumn{1}{|l|}{}                                                                                                                           & Q21. I learned strategies to cope with my problems.                                                                                & 4.50 $\pm$ 0.58
                      &                       \\ \cline{2-3}
\multicolumn{1}{|l|}{}                                                                                                                           & Q22.  I strengthened one or more self-management skills (i.e. time management).                                                                                                         & 4.50 $\pm$ 0.58
                      &                       \\ \cline{2-3}
\multicolumn{1}{|l|}{}                                                                                                                           & Q23. If given the chance, I would continue further sessions with the robot.                                                                                                 & 4.50 $\pm$ 1.00
                      &                       \\ \hline

\end{tabular}
\end{table*}